\documentclass[preprint,superscriptaddress,nofootinbib,preprintnumbers,amsmath,amssymb]{revtex4-1}

\usepackage{bbm}
\usepackage{stmaryrd}
\usepackage[bb=boondox]{mathalfa}
\usepackage{amsfonts}
\usepackage{mathrsfs}
\usepackage{leftidx}
\usepackage{amssymb}
\usepackage{placeins}
\usepackage{relsize}
\usepackage{slashed}
\usepackage{multirow}
\usepackage[dvipsnames]{xcolor}
\usepackage[colorlinks]{hyperref}
\usepackage{float}
\usepackage{natbib}

\usepackage{adjustbox}

\newcommand{\mymatrix}[1]{\ensuremath{\left\downarrow\vphantom{#1}\right.\overset{\xrightarrow[\hphantom{#1}]{ \text{\normalsize $\dot{\mu}$}}}{#1}}}
\makeatletter
\newcommand*{\ovA}[1]{%
  \m@th\overline{\mbox{$#1$}\raisebox{2.25mm}{}}%
}
\newcommand*{\ovB}[1]{%
  \m@th\overline{\mbox{$#1$}\raisebox{2.5mm}{}}%
}
\newcommand{\overleftrightsmallarrow}{\mathpalette{\overarrowsmall@\leftrightarrowfill@}}
\newcommand{\overrightsmallarrow}{\mathpalette{\overarrowsmall@\rightarrowfill@}}
\newcommand{\overleftsmallarrow}{\mathpalette{\overarrowsmall@\leftarrowfill@}}
\newcommand{\overarrowsmall@}[3]{%
  \vbox{%
    \ialign{%
      ##\crcr
      #1{\smaller@style{#2}}\crcr
      \noalign{\nointerlineskip}%
      $\m@th\hfil#2#3\hfil$\crcr
    }%
  }%
}
\def\smaller@style#1{%
  \ifx#1\displaystyle\scriptstyle\else
    \ifx#1\textstyle\scriptstyle\else
      \scriptscriptstyle
    \fi
  \fi
}

\makeatother

\def\be{\begin{equation}}
\def\ee{\end{equation}}
\def\bea{\begin{align}}
\def\eea{\end{align}}

\begin{document}

\newenvironment{psmallmatrix}
  {\left(\begin{smallmatrix}}
  {\end{smallmatrix}\right)}


\title{Moment-Matching Graph-Networks\\for Causal Inference\vspace{0.5cm}}
\author{Michael Park}
\email{q1park@gmail.com}
\affiliation{R\={o}nin, New York, NY\vspace{1cm}}  

\begin{abstract}
\vspace{0.25cm}
In this note we explore a fully unsupervised deep-learning framework for simulating non-linear structural equation models from observational training data. The main contribution of this note is an architecture for applying moment-matching loss functions to the edges of a causal Bayesian graph, resulting in a {\it generative conditional-moment-matching graph-neural-network}. This framework thus enables automated sampling of latent space conditional probability distributions for various graphical interventions, and is capable of generating out-of-sample interventional probabilities that are often faithful to the ground truth distributions well beyond the range contained in the training set. These methods could in principle be used in conjunction with any existing autoencoder that produces a latent space representation containing causal graph structures.
\end{abstract}

\maketitle

\section{Introduction}

Recently there have been many efforts to imbue deep-learning models with the ability to perform causal inference. This has been motivated primarily by the inability of traditional correlative models to make predictions on interventional and counterfactual questions \cite{pcrbook, pearlbook}, as well as the explainability of causal graphical models. These efforts have largely run in parallel to the developing trend of exploiting the non-local properties of graph neural networks \cite{DBLP:journals/corr/abs-1711-07971} to generate powerful and efficient representations of high-dimensional data.

\begin{figure}[ht]
  \begin{center}
    \includegraphics[scale=0.5]{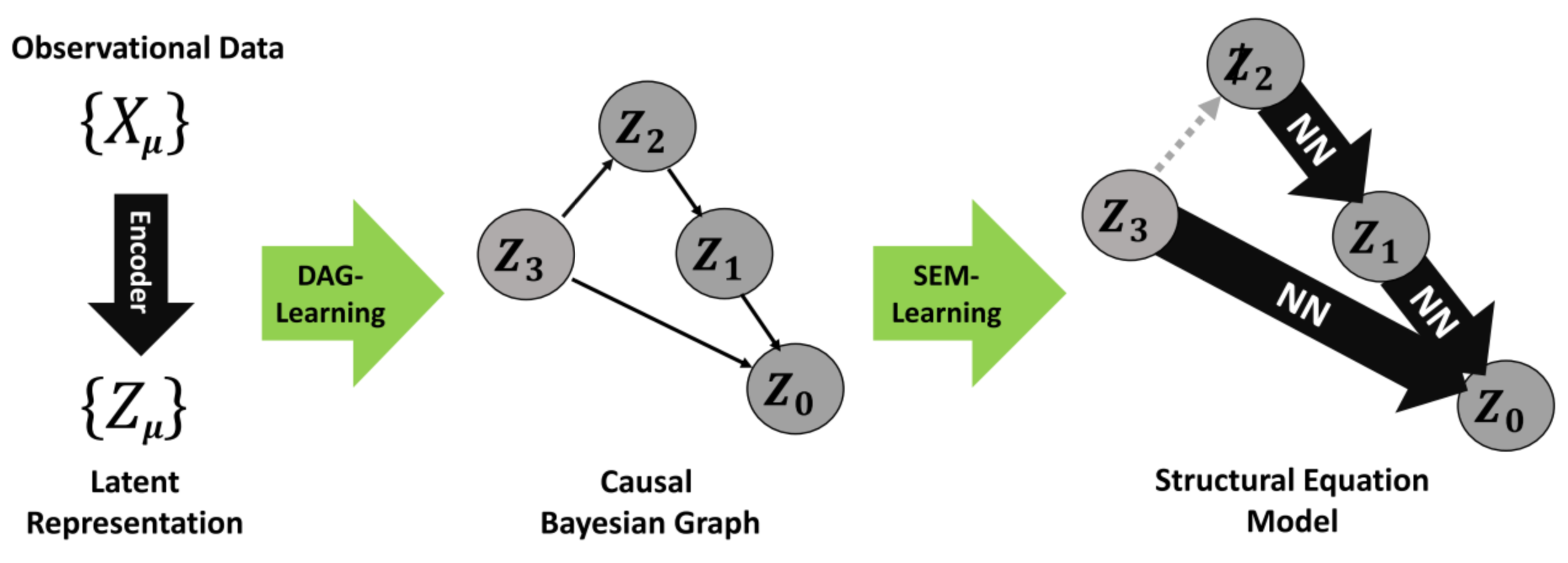}\vspace{-0.75cm}
  \end{center}
  \caption{The causal inference steps in this note begin with existing DAG structure-learning algorithms to infer causal structures in latent representations of data. Using the learned DAG, neural-networks are used to estimate the response of conditional probabilities under various graphical interventions.}
  \label{fig:2step}
\end{figure}

In this note we dichotomize the task of causal inference as a two-step process, illustrated in Figure \ref{fig:2step}. The first step involves inferring the graphical structure of a causal model associated with a given observational data set as a directed acyclic graph (DAG). Inferring the structure of causal DAG's from observational data has a long history and there have been many proposed techniques including constraint-based \cite{pcrbook, pearlbook, Zhang2008-ZHAOTC-3, 10.5555/2074158.2074204} and score-based methods \cite{10.1007/BFb0028180, Chickering2002OptimalSI, DBLP:journals/corr/abs-1302-3567, heckarticle}, recently developed masked-gradient methods \cite{zheng2018dags, zheng2019learning, DBLP:journals/corr/abs-1904-10098, ng2019graph, ng2019masked, fang2020low, ng2020role}, as well as hybrid methods \cite{DBLP:journals/corr/abs-1906-02226}. Notable novel alternatives also include methods based on reinforcement-learning \cite{DBLP:journals/corr/abs-1906-04477}, adversarial networks \cite{kalainathan2018structural} and restricted Boltzmann machines \cite{Sokolovska2020UsingUD}. Since the task of causal structural discovery is merely a means to an end for this work, we (rather arbitrarily) adopt the masked-gradient approach due to its parsimonious integration with the neural network based architectures for SEM-learning that are the subject of this note.\footnote{codebase: \url{http://github.com/q1park/spacetime}}

For the second step of causal inference, we develop a novel autoencoding architecture that applies generative moment-matching neural-networks \cite{DBLP:journals/corr/ZhaoSE17b, DBLP:journals/corr/RenLLZ16} to the edges of the learned causal graph, in order to estimate the functional dependence of the causally related observables as a structural equation model (SEM). Since their inception, generative moment-matching networks have been used for various tasks \cite{diane2017a, gaoproceed, briol2019statistical, lotfollahi2019conditional} related to the estimation of joint and conditional probability distributions, but to our knowledge this is the first use of their applications to an explicit causal graph structure. Our aim is to develop a fully unsupervised formalism that starts from purely observational tabular data, and ends with a robust automated sampling procedure that generates an accurate functional estimate of conditional probability distributions for the associated SEM. Existing techniques for Bayesian sampling on the latent space of generative models are also numerous, including Monte Carlo and gradient-optimization based methods \cite{ahn2012bayesian, 2001SPIE.4322..456H, DBLP:journals/corr/abs-1812-03285}. 

Much of this work has been inspired by several recent efforts to develop generative models that encode causal structure. For example, in \cite{DBLP:journals/corr/abs-1709-02023} the authors develop specific conditional adversarial loss functions for learning multi-step causal relations. Their goals are similar to those described in this note with a focus on linear relations within high-dimensional image vectors. In \cite{yang2020causalvae} the authors use supervised learning to endow the latent space distributions of a variational autoencoder with a causal graphical structure, with the aim of intervening on this latent space to control specific properties of their feature maps. In this note we perform experiments on simple low-dimensional feature maps, and examine the performance of our autoencoder in generating accurate conditional probability distributions from complex non-linear multi-step causal structures. These causal structures are assumed to exist as relations among dimensions in the latent representation of the data. Thus in principle, the methods described here should also be applicable to more complex feature maps such as those generated by image and language data. However experimentation on these high-dimensional data types are beyond the scope of this note.

In Section \ref{sec:bkg} we give a brief review of causal graphs and describe a vectorized formulation for structural equation models that is suited for deep-learning applications. In Section \ref{sec:exp} we give the results of our experiments on causal structure learning using existing masked gradient methods. We then describe our algorithm for SEM-learning and provide results on its performance. In Section \ref{sec:disc} we conclude with a discussion on possible applications and future directions for this work. 

\raggedbottom
\section{Background}
\label{sec:bkg}

\subsection{Causal Graphs}

\begin{figure}[ht]
  \begin{center}
    \includegraphics[scale=0.45]{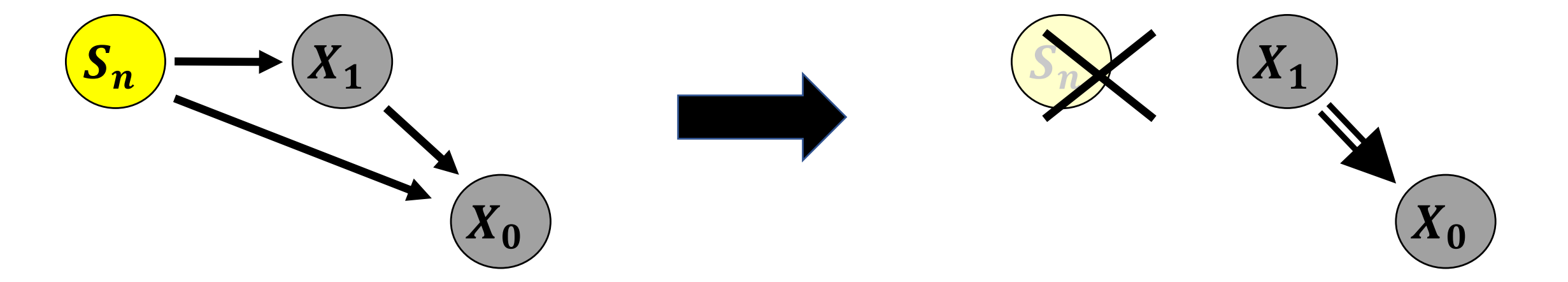} \vspace{-0.75cm}
  \end{center}
  \caption{Integrating out a confounding common cause variable $S_n$ generates a spurious correlation via a correction to the conditional probability distribution $P(X_0 | X_1)$.}
  \label{fig:spurion}
\end{figure}

\begin{figure}[ht]
  \begin{center}
  \includegraphics[scale=0.45]{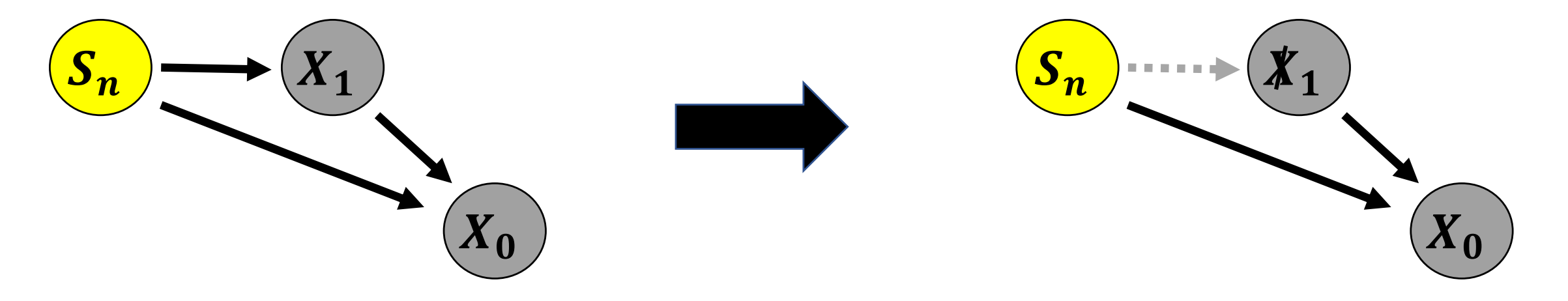} \vspace{-0.75cm}
  \end{center}
  \caption{Observing a controlled change to some variable $X_1$ requires removing the effects of any possible external influences. This is represented graphically by mutilating all in-going arrows into node $X_1$.}
  \label{fig:intervention}
\end{figure}

The identification of a causal effect between two variables is equivalent to measuring the response $\delta_0$ of some endogenous variable $X_0$ with respect to a controlled change $\delta_1$ in some exogenous variable $X_1$. If all of the variables are controlled, then the causal effect can be directly inferred via the conditional probability distribution $P(X_0 +\delta_0 | X_1+\delta_1)$. Inferring causal effects from uncontrolled observational data is challenging due to the existence of confounding variables $S_n$ which generate spurious correlations whose effects on the conditional probability $P(X_0 (S_n) | X_1 (S_n))$ may be statistically indistinguishable from true causal effects. This is illustrated diagramatically in Figure \ref{fig:spurion}. Here we adopt the formalism of Pearl in which the effect of a controlled change in variable $X_1$ is represented on a causal graph by mutilating all of the arrows going into node $X_1$ as shown in Figure \ref{fig:intervention}. The result is referred to as the {\it intervened}\footnote{For notational simplicity we use slashes to indicate graph mutilated variables in conditional probabilities rather than Pearl's original notation of $P(X_0|{\rm do}(X_1))$} conditional probability distribution $P(X_0|\slashed{X}_1) \sim P(X_0 +\delta_0 | X_1+\delta_1)$

\begin{figure}[ht]
  \begin{center}
  \includegraphics[scale=0.45]{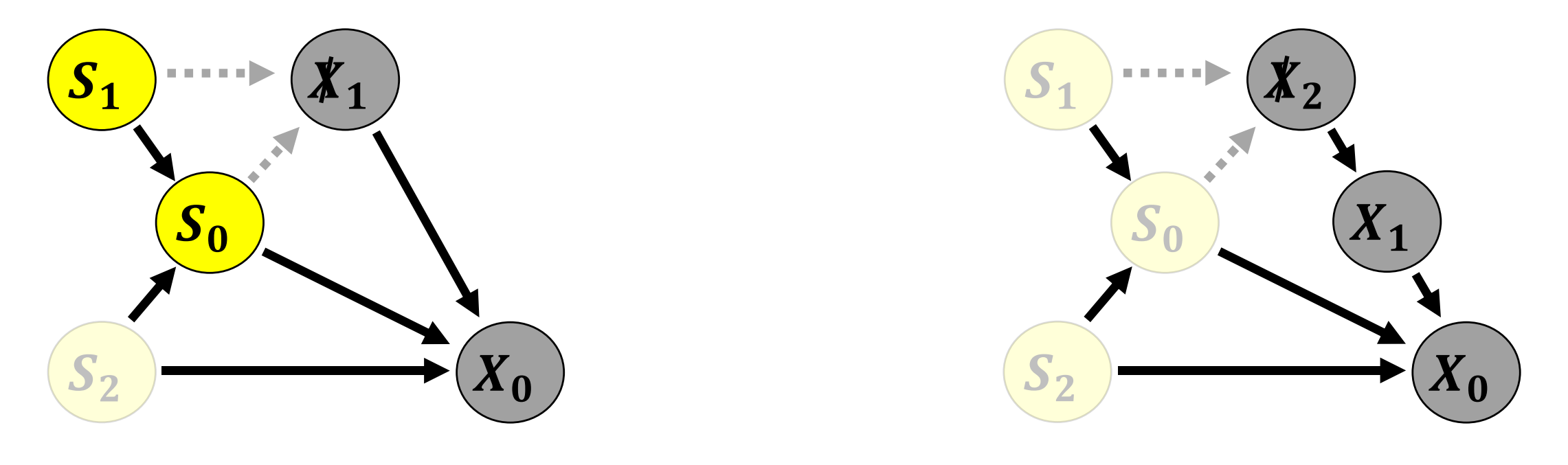} \vspace{-0.75cm}
  \end{center}
  \caption{(Left) Given the sufficiently complete set of measured confounding variables $S = \{ S_0, S_1 \}$, the back-door adjustment formula estimates the causal effect of $X_1$ on $X_0$. A measurement of only the set $S = \{ S_0 \}$ would be insufficient due to the existence of an unblocked ``back-door" path between the observables given by $X_1 \rightarrow S_1 \rightarrow S_0 \rightarrow S_2 \rightarrow X_0$. (Right) If there exists a causal chain $X_2 \rightarrow X_1 \rightarrow X_0$, the front-door adjustment formula can be used to disentangle the causal effect of $X_2$ on $X_0$ from any measured or unmeasured confounding variables.}
  \label{fig:backfront}
\end{figure}

There exists a rich literature describing the necessary and sufficient conditions for statistical distinguishability between causal and correlative effects, as well as methods for estimating causal responses when these conditions are met \cite{pcrbook, pearlbook}. Although the necessary conditions are beyond the scope of this brief review, the sufficient conditions amount to a requirement that the subset of measured confounding variables must be {\it sufficiently complete} so as to provide adequate control over the causal effects. In particular, the requirement of {\it sufficient completeness} can be succinctly dichotomized into two cases known as the {\it back-door} and {\it front-door} criterion. The {\it back-door criteria} can be used to estimate the causal response on a pair of nodes $X_1 \rightarrow X_0$, given an observation of a set of confounding variables $S = \{ S_0, S_1 \}$ as shown in Figure \ref{fig:backfront}. The intervened conditional probability can then be computed via the back-door adjustment formula given in Equation \ref{eq:adjustback}.
\begin{align}
    P(X_i | \slashed{X}_j = x) &= \displaystyle\int d s \, P(X_i | X_j = x, S=s) \, P(S=s) \label{eq:adjustback}
\end{align}
The {\it front-door criteria} can be used to estimate the causal response on a pair of nodes $X_2 \rightarrow X_0$ in situations where there exists a chain of causal influences $X_2 \rightarrow X_1 \rightarrow X_0$ as shown in Figure \ref{fig:backfront}. The intervened conditional probability can then be computed via the front-door adjustment formula given in Equation \ref{eq:adjustfront}.
\begin{align}
    P(X_i | \slashed{X}_j = x) &= \displaystyle\int d s \, P(S=s | X_j = x) \displaystyle\int d x^\prime \, P(X_i | X_j = x^\prime, S=s) \, P(X_j = x^\prime) \label{eq:adjustfront}
\end{align}

\subsection{Structural Equation Models}

\begin{figure}[ht]
  \begin{center}
  \includegraphics[scale=0.45]{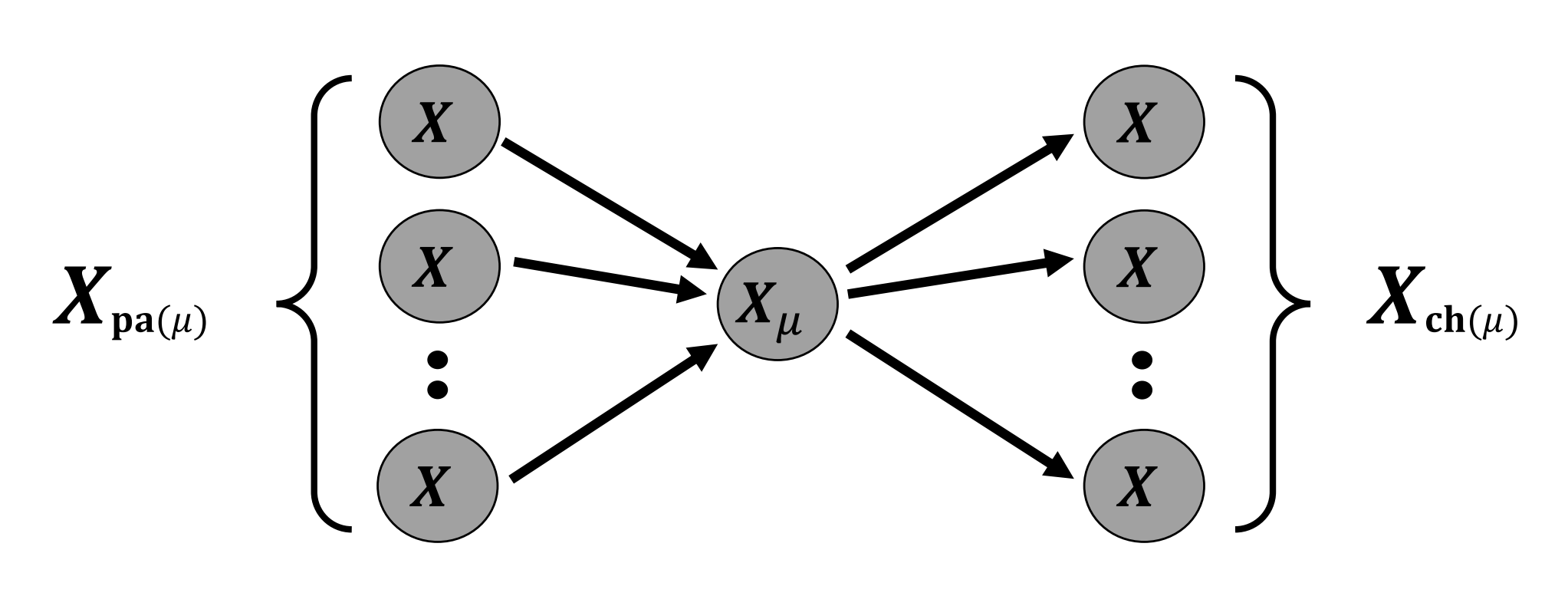} \vspace{-0.75cm}
  \end{center}
  \caption{Given some node in a causal graph $X_\mu$, we use $X_{{\rm pa}(\mu)}$ to refer to the set of all nodes that are parents of node $\mu$ and $X_{{\rm ch}(\mu)}$ to refer to the set of all nodes that are children of node $\mu$.}
  \label{fig:pach}
\end{figure}

Structural equation models (SEM's) are a functional extension of causal graphical models in which the values of each node variable $X_{\mu}$ are determined as a function of its parent node variables $X_{{\rm pa}(\mu)}$ and noise $\xi_\mu$. Here we adopt a notation where each node in a causal graph with $V$ nodes is specified by a spacetime index $\mu = 1, ..., V$ and Einstein summation is assumed. The set of parent (child) nodes corresponding to $\mu$ is given by $X_{{\rm pa}(\mu)}$ ($X_{{\rm ch}(\mu)}$) as illustrated in Figure \ref{fig:pach}. The generic form for an SEM can then be expressed as shown in Equation \ref{eq:sem}
\begin{equation}
    X_{\mu} = f \left( \xi_\mu, \, X_{ {\rm pa} (\mu) } \right)
    \label{eq:sem}
\end{equation}

If the contribution from noise is assumed to be additive, then each node variable $X_\mu$ can be expressed simply as a polynomial (or other) expansion in its parent nodes $X_{{\rm pa} (\mu)}$ as shown in Equation \ref{eq:polysem}. The leading order term in this expansion describes a linearized SEM, which is typically expressed in terms of a weighted graph adjacency matrix $W_{\mu \nu}$ in the form shown in Equation \ref{eq:linsem}.
\begin{align}
    X_\mu &= -\xi_\mu + f \left( X_{ {\rm pa} (\mu) } \right) \nonumber \\
    &\approx - \xi_\mu + \displaystyle\sum_{n=1}^\infty c_{n,{\rm pa} (\mu) } X_{ {\rm pa} (\mu) }^n \label{eq:polysem} \\
    &\xrightarrow{\mathcal{O}(1)} - \xi_\mu + W_{\mu \nu} X_\nu \label{eq:linsem}
\end{align}
 The linear SEM of Equation \ref{eq:linsem} has the unique property that its exact solution describes a generative model that predicts each variable from pure noise as shown in Equation \ref{eq:gensem}. The inverse operator can be expressed in closed-form as a degree-$d$ polynomial in terms of Cayley-Hamilton coefficients $c_n$, which describe the propagation of ancestral noise through the causal graph. Thus each node variable $X_\mu$ can be expressed as a linear combination of its noise $\xi_\mu$ and the noise of its $n^{\rm th}$ ancestors $\xi_{{\rm pa}_n (\mu)}$, as shown in Equation \ref{eq:noiseprop}.
\begin{align}
    X_\mu &= \left( - \delta_{\mu \nu} + W_{\mu \nu} \right)^{-1} \xi_\nu \label{eq:gensem} \\
    &= \left( - \delta_{\mu \nu} + \displaystyle\sum_{n=1}^d c_n W_{\mu \nu}^n \right) \xi_\nu \nonumber \\
    &= - \xi_\mu + \displaystyle\sum_{n=1}^d c_n \, \xi_{{\rm pa}_n (\mu)} \label{eq:noiseprop}
\end{align}

The weighted adjacency matrix $W_{\mu \nu}$ serves the dual purpose of masking each node variable $X_\mu$ from its non-parent nodes through its zero-entries, while the non-zero entries define the strength of linear correlations between each pair of nodes in the causal graph. Unfortunately there is no standardized generalization to non-linear SEM's. One natural possibility is to define a separate weighted adjacency matrix $W_{\mu \nu}^{(n)}$ for each order $n$ in a functional expansion like the polynomial example in Equation \ref{eq:polysem}. While this interpretation nicely generalizes the linear approximation, its computational complexity is unbounded, and there have been various other suggested interpretations for the adjacency matrix weights, related to the mutual information between parent-child node variables \cite{fang2020low}. 

In this note we develop an alternative formalism for describing non-linear SEM's that is agnostic to the interpretation of the weights in the adjacency matrix. We thus define a causal mask matrix $M_{\mu \nu}$ which is just the unweighted adjacency matrix as shown in Equation \ref{eq:maskmatrix}, where $\odot$ refers to an element-wise multiplication. 
\begin{align}
    M_{\mu \nu} \equiv | W_{\mu \nu} | \odot \frac{1}{|W_{\mu \nu}| + \epsilon} \label{eq:maskmatrix}
\end{align}
We then define a procedure for extracting the data for the parents of each node in the following way. We first lift each node variable into an auxiliary dimension $\dot{\mu} = 1, ..., V$. Index contraction of the spacetime index with the mask matrix $M_{\mu \nu}$ then produces a vector $X_{{\rm pa} (\mu)}^{\dot{\mu}}$ for each node $\mu$ whose index in the auxiliary dimension contains its parent-node data as shown in Equation \ref{eq:nodehot}. This vectorized parental masking procedure is suitable for expressing functions of sets of parent-nodes in a generalized SEM as $X_{\mu}^{\dot{\mu}} = f ( \xi_\mu, \, X_{ {\rm pa} (\mu) }^{\dot{\mu}} )$.
\begin{align}
    X_\mu &~\longrightarrow~ X_\mu^{\dot{\mu}} \equiv X_\mu \otimes \delta_\mu^{\dot{\mu}} = \quad \text{\normalsize $\mu$}\mymatrix{
    \begin{pmatrix}
    X_V & 0 & \cdots & 0 & 0 \\
    0 & X_{V-1} & \cdots & 0 & 0 \\
    \vdots  & \vdots  & \ddots & \vdots & \vdots \\
    0 & 0 & \cdots & X_1 & 0 \\
    0 & 0 & \cdots & 0 & X_0
    \end{pmatrix}
    } \nonumber \\
    &~\longrightarrow~ M_{\mu \nu} X_\nu^{\dot{\mu}} = \begin{pmatrix}
    0 & 0 & \cdots & 0 & 0 \\
    X_V & 0 & \cdots & 0 & 0 \\
    \vdots  & \vdots  & \ddots & \vdots & \vdots \\
    X_V & X_{V-1} & \cdots & 0 & 0 \\
    X_V & X_{V-1} & \cdots & X_{1} & 0 
    \end{pmatrix} = \begin{pmatrix}
    X_{{\rm pa} (V)}^{\dot{\mu}} \\
    ~~\ X_{{\rm pa} (V-1)}^{\dot{\mu}} \\
    \vdots  \\
    X_{{\rm pa} (1)}^{\dot{\mu}} \\
    X_{{\rm pa} (0)}^{\dot{\mu}}
    \end{pmatrix} = X_{{\rm pa} (\mu)}^{\dot{\mu}} \label{eq:nodehot}
\end{align}

\section{Experiments}
\label{sec:exp}

\subsection{Causal Structure Learning}

The algorithms for SEM-learning described in this note rely on first inferring the correct causal graph structure for a given data set. Fortunately the last two years have seen exciting progress in applications of neural networks to the problem of causal graph structure-learning, particularly in the area of masked-gradient methods \cite{zheng2018dags, DBLP:journals/corr/abs-1904-10098, ng2019graph, fang2020low, ng2020role}. These methods center around an identity for acyclic weighted adjacency matrices, which was first derived in \cite{zheng2018dags} and is shown in Equation \ref{eq:acyclic}. This identity enables a re-formulation of acyclic graph-learning as a continuous optimization problem. Here again $\odot$ denotes element-wise multiplication.
\begin{align}
    {\rm tr} \, e^{W \odot W} = {\rm tr} \, I \label{eq:acyclic}
\end{align}
The graph-learning network can then be constructed using an encoder/decoder framework with an objective function that attempts to minimize some reconstruction loss, subject to an acyclicity constraint $h=0$, where $h$ is a function of the weighted adjacency matrix given in Equation \ref{eq:acconstraint}.
\begin{align}
    h(W) = - {\rm tr} \, I + {\rm tr} \, e^{W \odot W} = 0 \label{eq:acconstraint}
\end{align}
The original formulation for this continuous optimization, referred to as $\texttt{NO-TEARS}$ \cite{zheng2018dags}, uses a reconstruction loss inspired directly by the form of the linear SEM in Equation \ref{eq:linsem}. As illustrated in in the first line of Table \ref{tab:structalgos}, the encoder $\mathcal{E}$ is just the identity function while the decoder $\mathcal{D}$ is an MLP that takes as input a weighted masked latent space vector $W \cdot Z$.

\bgroup
\def\arraystretch{1.5}
\begin{table}[ht]
    \begin{tabular}{ccc}
     & Encoder & Decoder \\
     \hline
    \texttt{NO-TEARS}: & \qquad $Z = X$ \qquad & \qquad $\widehat{X} = \mathcal{D}(W \cdot Z)$ \\
    \texttt{GNN}: & \qquad $Z = (-I+W) \cdot \mathcal{E} (X)$ \qquad & \qquad $\widehat{X} = \mathcal{D}((-I+W)^{-1} \cdot Z)$ \\
    \texttt{GAE}: & \qquad $Z = \mathcal{E}(X)$ \qquad & \qquad $\widehat{X} = \mathcal{D} ( W \cdot Z)$
    \end{tabular} 
    \caption{A comparison of functional structures for three well known masked-gradient-based algorithsm for causal structure learning.}
    \label{tab:structalgos}
\end{table}
\egroup

In this note we focus our tests on two non-linear generalizations of the $\texttt{NO-TEARS}$ algorithm, referred to as $\texttt{GNN}$ and $\texttt{GAE}$. The encoder/decoder architectures are given in Table \ref{tab:structalgos}, where $\mathcal{E}$ and $\mathcal{D}$ refer to generic MLP based function-learners. Both of the $\texttt{GNN}$ and $\texttt{GAE}$ frameworks generalize the well known closed-form solution for linear SEM's. However the salient difference between them is the presence of a residual connection in \texttt{GNN} represented by the identity term in the second line of Table \ref{tab:structalgos}. The reconstruction loss function for $\texttt{GNN}$ is given by the usual evidence lower-bound (ELBO) for variational autoencoders while the reconstruction loss for $\texttt{GAE}$ is simply the mean-squared-error (MSE). The above optimization can be implemented using the method of Lagrange multipliers with the Lagrangian defined in Equation \ref{eq:lagrangian}.

\begin{align}
    \mathcal{L}_\texttt{GNN/GAE} &= -\mathcal{L}_{\rm ELBO/MSE} + \lambda \, | h(W_{\mu \nu}) | + \frac{c}{2} \, | h(W_{\mu \nu}) |^2 \label{eq:lagrangian}
\end{align}

Following the work in \cite{DBLP:journals/corr/abs-1904-10098, ng2019graph} we perform tests on four different toy data sets generated by structural equation models of increasing non-linear complexity, as shown in Equations \ref{eq:egsem1}-\ref{eq:egsem4}.
\begin{align}
    \text{linear:}& \quad X = -\xi + W \cdot X \label{eq:egsem1} \\
    \text{non-linear 1:}& \quad X = -\xi + W \cdot  \cos \ (X + 1) \label{eq:egsem2} \\
    \text{non-linear 2:}& \quad X = -\xi + 2 \, \sin \left( W \cdot (X + 1/2) \right) + W \cdot (X + 1/2) \label{eq:egsem3} \\
    \text{non-linear 3:}& \quad X = -\xi + 2 \, \sin \left( W \cdot ( \cos \ (X + 1) + 1/2) \right) + W \cdot ( \cos \ (X + 1) + 1/2) \label{eq:egsem4}
\end{align}
In the original papers, both $\texttt{GNN}$ and $\texttt{GAE}$ were tested using randomly generated Erd\H os-R\'enyi graphs. For graphs with $V$ nodes, the authors of $\texttt{GNN}$ reported structural hamming distance (SHD) errors ranging from $0.2 \times V$ (for nonlinear 2) and $0.8 \times V$ for (nonlinear 1). Impressively, the performance of the $\texttt{GAE}$ algorithm exhibits a scaling that is roughly independent of the number of nodes in the graph for the Erd\H os-R\'enyi case, which we have verified in our own experiments. The primary reason for the difference in performance on large graphs is due to the presence of the residual connection in $\texttt{GNN}$, which enables an extremely accurate reconstruction of the data despite an incorrect causal graph structure. 

\begin{figure}[ht]
  \begin{center}
    \includegraphics[scale=0.45]{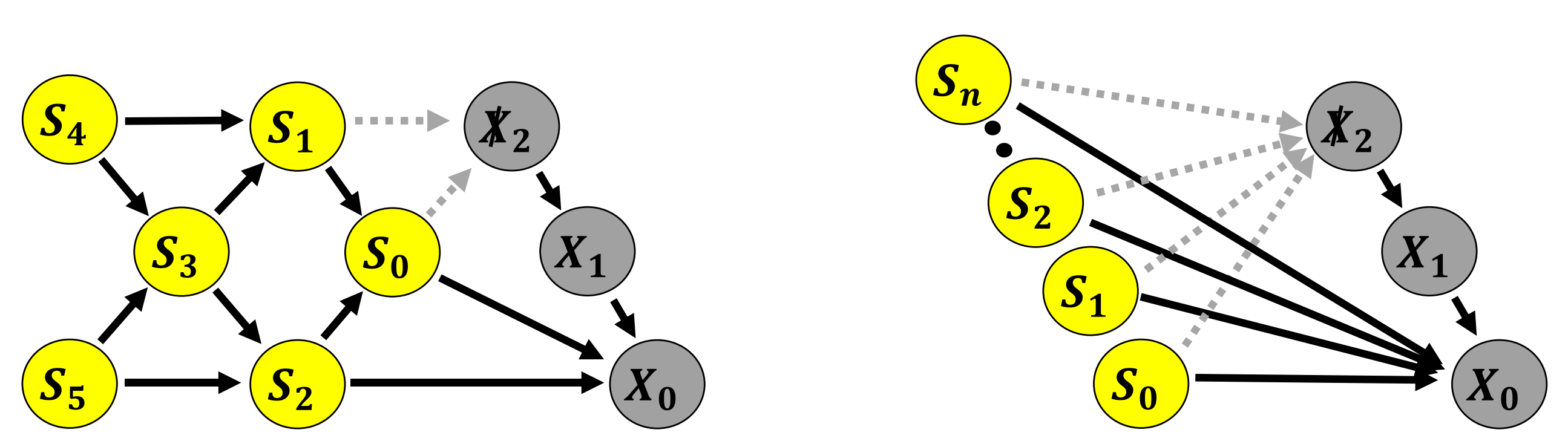}\vspace{-0.5cm}
  \caption{Two graph structures used for the experiments in this note, which we refer to as Graph A (left) and Graph B (right). Causal estimation for Graph A requires mutilating two edges independent on the number of confounders, while causal estimation for Graph B requires mutilating a number of edges equal to the number of confounders.}
  \label{fig:graph}
  \end{center}
\end{figure}

\begin{figure}[ht]
  \begin{center}
    \includegraphics[scale=0.52]{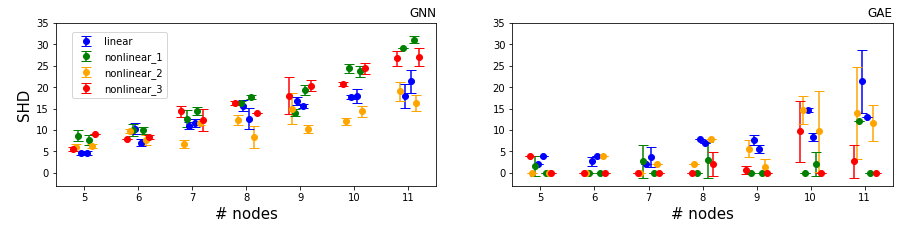}
    \includegraphics[scale=0.52]{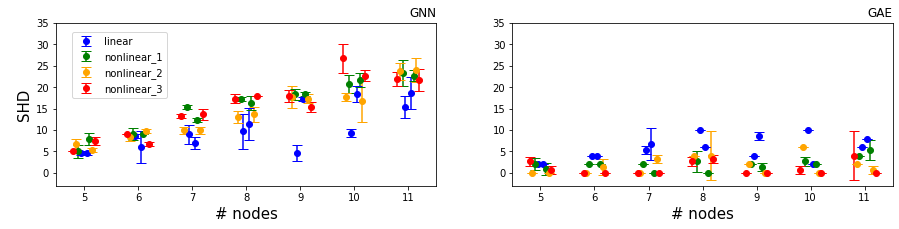}\vspace{-0.5cm}
  \end{center}
  \caption{Structural hamming distances (SHD) for \texttt{GNN} and \texttt{GAE} as a function of the total number of nodes. Results are shown for Graph A (top row) and Graph B (bottom row) as defined in \ref{fig:graph}. For each \# nodes we generate two graphs with different weights from different random seeds and perform 3 runs for each graph. The error bars indicate variations between the 3 runs on each seed.}
  \label{fig:shd}
\end{figure}

In this note we perform tests on the $\texttt{GNN}$ and $\texttt{GAE}$ algorithms using the two graph structures shown in Figure \ref{fig:graph}, referred to as Graph A and Graph B. These two graph structures form the baseline cases for our structural equation model tests described in the next section, and represent different configurations of confounding variables increasing in number. The results of our structure-learning experiments, shown in Figure \ref{fig:shd}, indicate that the explicit presence of numerous confounding variables presents a significant obstacle to the recovery of correct causal structures relative to the Erd\H os-R\'enyi case, even for simple graphs with nodes as few as $\mathcal{O}(10)$. 

\subsection{Structural Equation Modeling}

\begin{figure}[ht]
  \begin{center}
    \includegraphics[scale=0.4]{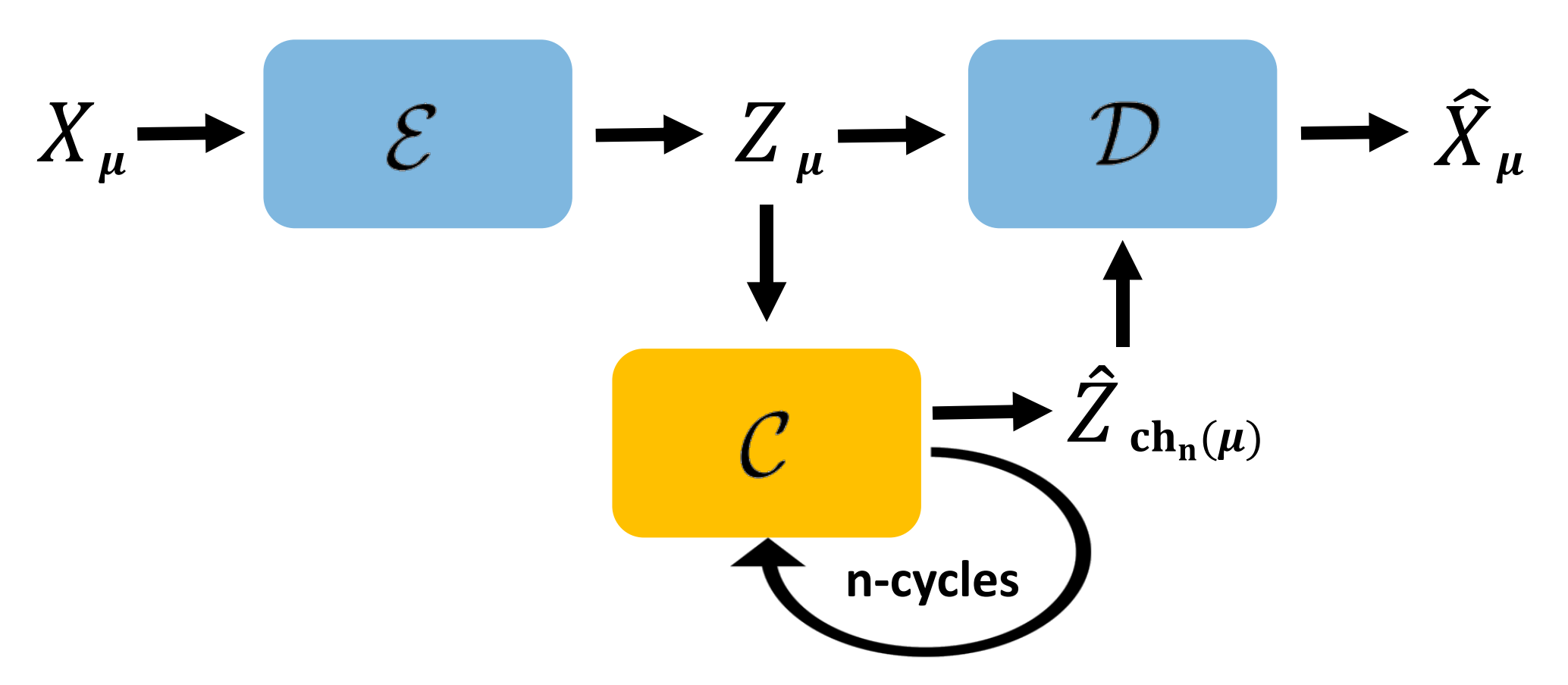}\vspace{-0.75cm}
  \end{center}
  \caption{The proposed network architecture is an extension of a generic variational autoencoder (blue). The generator for the latent space $Z_\mu$ is augmented with an additional causal network block $\mathcal{C}$ (orange) that uses a causal mask $M_{\mu \nu}$ as defined in \ref{eq:maskmatrix} to generate a latent space distribution for each child node $\widehat{Z}_{{\rm ch} (\mu)}$ that is a function of only it's parent nodes $Z_\mu$. The $n^{th}$ child node of a latent variable $Z_\mu$ can thus be generated by cycling the inputs $n$ times through $\mathcal{C}$.}
  \label{fig:archi}
\end{figure}

The network architecture for SEM-learning proposed in this note is illustrated in Figure \ref{fig:archi}, and can be factorized into two components. The first component is just a generic variational autoencoder that encodes each node feature $X_\mu$ into its latent representation $Z_\mu$ before decoding it back to the target representation $\widehat{X}_\mu$. The second component introduces a ``causal block" $\mathcal{C}$ that performs ancestral sampling on the latent representation $Z_\mu$ and produces a latent representation for each child-node $\widehat{Z}_{{\rm ch} (\mu)}$ that is a function of \textit{only its parent-nodes} $Z_\mu$. 

\begin{figure}[ht]
  \begin{center}
    \includegraphics[scale=0.4]{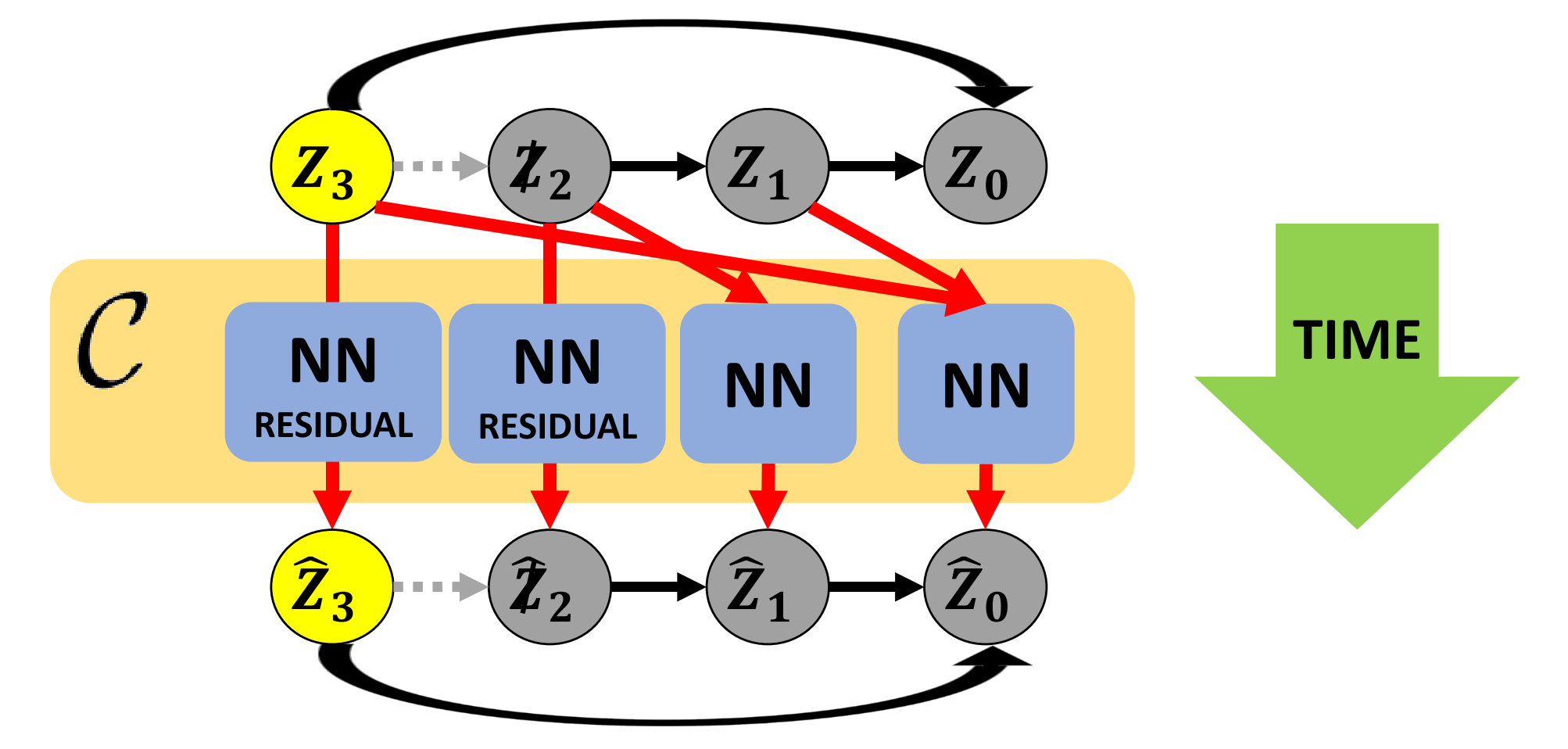}\vspace{-0.75cm}
  \end{center}
  \caption{The causal block $\mathcal{C}$ takes inputs from the latent node variables $Z_\mu$. A single neural network for each latent dimension generates means and variances for the child nodes $\widehat{Z}_\mu$. Nodes with no parents, including the intervened node $Z_2$, contain a residual connection, and all nodes with parents are functions of only their parents.}
  \label{fig:sampling}
\end{figure}

For SEM-learning on a graph with $V$ nodes, the causal block $\mathcal{C}$ is correspondingly composed of $V$ neural-networks as illustrated diagramatically in Figure \ref{fig:sampling}. A restriction on the functional dependence of each node to only its parent nodes is crucial for the automated generation of intervened conditional probability distributions. This is achieved simply through the use of the causal mask $M_{\mu \nu}$ in the causal block $\mathcal{C}$, as well as the absence of any residual connection except for those nodes which have no parents. This includes those nodes which are chosen for intervention, as well as those nodes with no parents since they can be viewed as being intervened on by the environment. Ancestral sampling of an intervened distribution can then be performed simply by generating data for the intervened node $Z_\mu$ from a random-normal distribution, and cycling the data through the causal block $n$ times in order to obtain the data for its $n^{\rm th}$ child node $Z_{{\rm ch_n} (\mu)}$, as illustrated in \ref{fig:archi}.

A functional expression for the causal block $\mathcal{C}$ can be expressed as a sum of three terms as shown in Equation \ref{eq:causalsem}. The first term $\xi_\mu$ describes the contribution from noise and is computed via the usual reparameterization trick \cite{kingma2013autoencoding} from neural-network-generated variances. The second term provides a residual connection only for node variables that have no parents. We thus define a delta function $\delta_{{\rm pa} (\mu)}$ whose argument given a specified node $\mu$ is the number of parents belonging to that node, and normalized as shown in equation \ref{eq:parentres}.
\begin{align}
\mathcal{C}(Z_\mu) &=  - \xi_\mu - \delta_{{\rm pa}(\mu)} Z_\mu + \left( 1-\delta_{{\rm pa}(\mu)} \right) {\rm NN}_\mu^{\dot{\mu}} ( Z_{{\rm pa} (\mu)}^{\dot{\mu}} ) \label{eq:causalsem} \\
&\longrightarrow \widehat{Z}_{{\rm ch} (\mu)} \nonumber
\end{align}

\begin{equation}
  \delta_{{\rm pa}(\mu)}=\left\{
  \begin{array}{@{}ll@{}}
    1 & ~~\text{if \# parents = 0 for node}\ \mu \\
    0 & ~~\text{otherwise}
  \end{array}\right. \label{eq:parentres}
\end{equation} 
The third and final term is generated by the set of $V$ neural networks ${\rm NN}_\mu^{\dot{\mu}}$ whose input is the vector containing the latent representation of $\mu$'s parent node data $Z_{{\rm pa} (\mu)}^{\dot{\mu}}$, as constructed according to Equation \ref{eq:nodehot}. The loss function used is a combination of the joint \cite{DBLP:journals/corr/ZhaoSE17b} and conditional \cite{DBLP:journals/corr/RenLLZ16} maximum-mean-discrepancies (MMD and CMMD) as shown in Equation \ref{eq:caeloss}, with $\gamma \gg \beta$. The set of networks ${\rm NN}_\mu^{\dot{\mu}}$ thus together form a generative conditional moment-matching graph-neural-network.

\begin{align}
    \mathcal{L} = &- \beta  \, D_{\rm MMD} \big( Q(Z|X) || P(Z) \big) - \gamma \, D_{\rm CMMD} \big( Q(\widehat{Z}|Z_{\rm pa}) || P(Z | Z_{\rm pa}) \big) \nonumber \\
    &+E_{Q(Z|X)} \big( \log P(\widehat{X} | Z ) \big) \label{eq:caeloss}
\end{align}

\begin{figure}[ht]
  \begin{center}
    \includegraphics[scale=0.5]{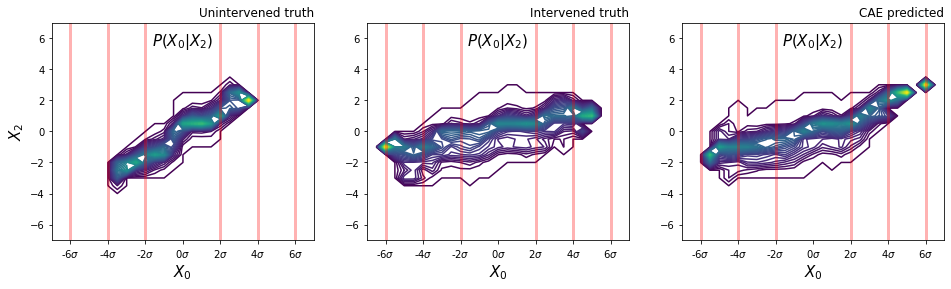}
    \includegraphics[scale=0.5]{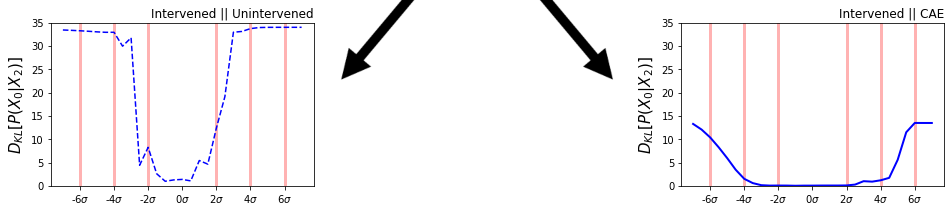}\vspace{-0.75cm}
  \end{center}
  \caption{The performance metric adopted in this note is the relative entropy $D_{KL}$ between the conditional probability distribution for the predicted intervened SEM (top right) and the ground truth SEM (top middle). The $D_{KL}$ is computed along slices corresponding to points at various standard deviations away from the mean (bottom right). As a baseline we compare this against the $D_{KL}$ with respect to the unintervened conditional probability distribution (bottom left).}
  \label{fig:metric} 
\end{figure}

To measure the performance of interventional sampling we perform tests using an MLP-based encoder and decoder $\mathcal{E}$/$\mathcal{D}$ each consisting of a single hidden layer with 16 neurons. The causal block $\mathcal{C}$ is composed of $V$ neural networks, each with input dimension $V$ and output dimension $1$, and each consisting of a single hidden-layer containing 64 neurons. For the loss function we choose (rather arbitrarily) $\beta=1$ and $\gamma=300$, and each trial is run on 8000 data points. The performance metric used is the relative entropy (KL divergence) between the conditional probability distributions generated by the intervened and unintervened ground truth SEM's $D_{\rm KL} \left( P(X_i | \slashed{X}_j = x_j) || Q (X_i | \slashed{X}_j = x_j) \right)$. We then compare it with the relative entropy between the intervened SEM and the one predicted by the causal autoencoder $D_{\rm KL} \left( P(X_i | \slashed{X}_j = x_j) || Q(X_i | X_j = x_j) \right)$ at different standard deviations away from the distribution means, as illustrated in Figure \ref{fig:metric}. The autoencoder predictions for these results have been smoothened using a kernel density estimator with a normal reference bandwidth.

\begin{figure}[ht]
  \begin{center}
    \includegraphics[scale=0.5]{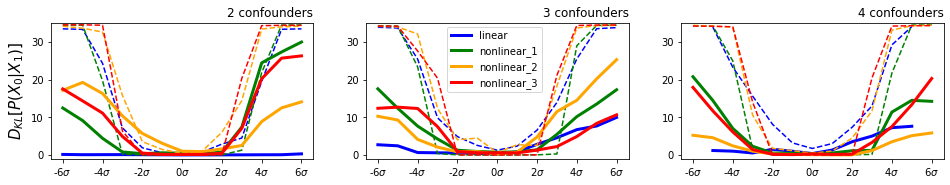}
    \includegraphics[scale=0.5]{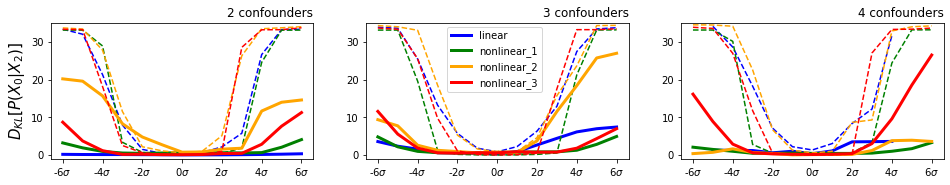}\vspace{-0.75cm}
  \end{center}
  \caption{Performance metrics for experiments on Graph A. $D_{KL}$'s are shown along contours of varying standard deviation $\sigma$ for the probability distributions $P(X_0 | X_1)$ (top row) and $P(X_0 | X_2)$ (bottom row). The solid and dashed lines represent averages for 4 randomly generated adjacency matrices.}
  \label{fig:resultsA}
\end{figure}
\begin{figure}[ht]
  \begin{center}
    \includegraphics[scale=0.5]{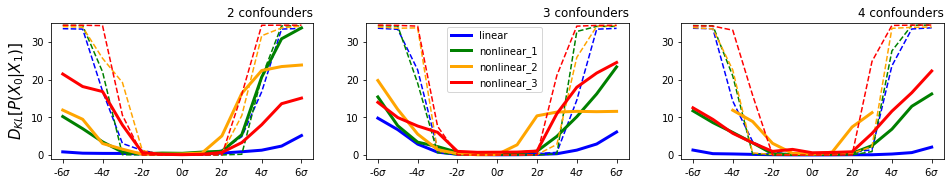}
    \includegraphics[scale=0.5]{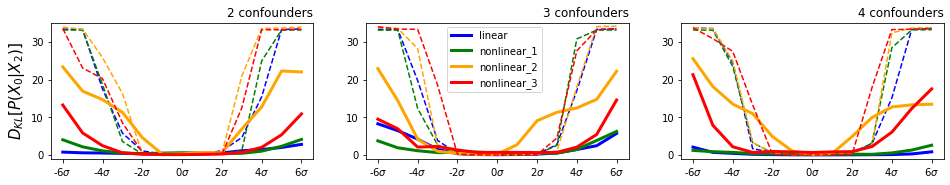}\vspace{-0.75cm}
  \end{center}
  \caption{Performance metrics for Graph B along contours of varying standard deviation $\sigma$. Results are shown for the probability distributions $P(X_0 | X_1)$ (top row) and $P(X_0 | X_2)$ (bottom row). The solid and dashed lines represent averages for 4 randomly generated adjacency matrices.}
  \label{fig:resultsB}
\end{figure}

\section{Discussion}
\label{sec:disc}

The results of our experiments indicate that the proposed framework for simulating structural equation models is capable of capturing complex non-linear relationships among variables in way that is amenable to multi-step counterfactual interventions. Importantly, the generated probability distributions appear faithful to the ground truth intervened SEM's, even when the intervened variables are fixed to values that are outside the range of values contained in the training data distributions. This capability implies a predictive ability that is manifestly beyond what is possible through analytical calculations via the back-door and front-door adjustment formulas, which can only be applied to intervened variables that take on values for which observable data exists.

With 8000 data points in each of the training sets, the maximum and minimum values for the node variable $X_2$ typically fall within the range of $3.5 \sigma$ from the distribution mean, never exceeding $4.0 \sigma$. From Figure \ref{fig:resultsA} and \ref{fig:resultsB}, we can observe that the linearly correlated data sets are faithful to the ground truth well beyond the $4.0 \sigma$ mark. On the other hand, those data sets with strong non-linear components vary in their predictive performance beyond $3 \sigma$, but are reliably closer to the ground truth relative to the un-intervened distributions. This is unsurprising upon closer inspection of the predicted conditional (intervened) probabilities, which demonstrate a clear tendency for our generative model to perform simple linear extrapolations of the distributions in regimes outside those contained in the training data.

Although the experiments performed in this note were restricted to the case of scalar-valued node variables, we expect that a very simple extension of these methods could make them applicable to complex high dimensional image and language data. For example in CausalVAE \cite{yang2020causalvae}, the authors use supervised learning to encode specific image labels into a single dimension of the latent space $Z_\mu$. In one example, they use the CelebA data set of facial images to encode causal relationships between features like $Age \rightarrow Beard$, thus allowing them to intervene on the latent space to produce images of unnaturally young bearded faces. Augmenting this procedure with the causal block $\mathcal{C}$ described in this note would in principle enable synthetic generation of image populations with features that accurately represent conditional probabilities under multiple steps of causal influence. For example, an accurate distribution of hair colors if the graph structure contained $Age \rightarrow Beard \rightarrow Hair \  Color$. Unfortunately a detailed exploration on these high dimensional data types is beyond the scope of this note.

Another potential application of these methods could be for use with model-based reinforcement learning. In \cite{DBLP:journals/corr/abs-1901-08162} the authors performed several experiments in a model-free RL framework in which they trained agents to make causal predictions in simple one-step-querying scenarios. In these experiments, the agents were directed to sample points from joint and conditional probability distributions of SEM-generated data, as well as the corresponding distributions from arbitrarily mutilated SEM graphs. These experiments showed evidence that their agents learned to exploit interventional and counterfactual reasoning to accumulate significantly higher rewards compared to the relevant baselines. 

In \cite{nair2019causal} the authors expand on the previous work by successfully training RL agents to perform causal reasoning in a more complex multi-step relational scenario with the ability to generalize to unseen causal structures that were held-out during training. Their experiments involved two separate RL agents. One which used supervised learning to generate a causal graph model off ground truth graphs, and another which was directed to take ``goal-oriented" actions based on models learned by the first agent. The authors strongly hypothesized that the impressive level of generalizability displayed by their algorithm was a direct result of the explicit model-based approach. We find the possibility of performing such experiments using graphical models learned via the fully unsupervised approach described in this note to be both very intriguing and plausibly practical as a future area of exploration.

\section{Acknowledgements}

We thank Vincent Tang, Jiheum Park, Ignavier Ng, Jungwoo Lee, and Tim Lou for useful discussions.

\bibliographystyle{unsrtnat}
\bibliography{main}

\end{document}